\documentclass{llncs}

\usepackage{amsmath}
\usepackage{amssymb}
\usepackage{mathrsfs}
\usepackage{mathtools}
\usepackage{listings}
\usepackage{graphicx}
\usepackage{caption}
\usepackage{subcaption}
\usepackage{wrapfig}

\usepackage{tikz}
\usetikzlibrary{arrows}
\usetikzlibrary{backgrounds}	% drawing the background after the foreground
\usetikzlibrary{automata,positioning}
\usetikzlibrary{decorations.pathmorphing} % To have curved arrows
\usetikzlibrary{shapes,snakes}
\usetikzlibrary{matrix,decorations.pathreplacing}

\usepackage{varwidth} %Per avere i blocchi di una lunghezza massima

\usepackage{hyperref}

\usepackage[ruled,lined]{algorithm2e}

\usepackage{multicol}
\usepackage{longtable}
\usepackage{array}

\usepackage[framemethod=tikz]{mdframed}

\usepackage{enumitem}

\usepackage[utf8]{inputenc}

\DeclareMathAlphabet{\mathcal}{OMS}{cmsy}{m}{n}

\usepackage{afterpage}
\usepackage{pdflscape}

\newcommand{\vars}[1]{\ensuremath{\mathit{Vars}(#1)}}
\newcommand{\adom}[1]{\ensuremath{\textsc{adom}(#1)}}
\newcommand{\ans}[1]{\ensuremath{\textsc{ans}(#1)}}

\newcommand{\cls}[1]{\ensuremath{\mathsf{#1}}}   % classes
\newcommand{\rol}[1]{\ensuremath{\mathsf{#1}}}   % relations
\newcommand{\cons}[1]{\ensuremath{\mathsf{#1}}}  % constants
\newcommand{\acname}[1]{\ensuremath{\mathsf{#1}}}
\newcommand{\acshort}[3]{\ensuremath{\acname{#1}}:~#2~$\rightsquigarrow$~#3}  % \acshort{Name}{Condition}{Effect}
\newcommand{\answ}{\ensuremath{\textsc{ans}}}

\newcommand{\asgn}{\ensuremath{:=}}

\mathchardef\breakingcomma\mathcode`\,
{\catcode`,=\active
  \gdef,{\breakingcomma\discretionary{}{}{}}
}

\title{Optimizations for Decision Making and Planning \\
in Description Logic Dynamic Knowledge Bases}

\author{Michele Stawowy}
\institute{IMT Institute for Advanced Studies, 
Lucca, Italy \\
\email{michele.stawowy@imtlucca.it}}

\begin{document}

\maketitle

\begin{abstract}
  Artifact-centric models for business processes recently raised a lot of attention, as they manage to combine structural (i.e. data related) with dynamical (i.e. process related) aspects in a seamless way.
Many frameworks developed under this approach, although, are not built explicitly for planning, one of the most prominent operations related to business processes.
In this paper, we try to overcome this by proposing a framework named Dynamic Knowledge Bases, aimed at describing rich business domains through Description Logic-based ontologies, and where a set of actions allows the system to evolve by modifying such ontologies.
This framework, by offering action rewriting and knowledge partialization, represents a viable and formal environment to develop decision making and planning techniques for DL-based artifact-centric business domains.
\end{abstract}

\section{Introduction}

Classically, management of business processes always focused on workflows and the actions/interactions that take part in them, an approach called \emph{process-centric}.
One of the most prominent operations related to business processes is \emph{planning} \cite{Nau:2004:APT:975615}, namely finding a sequence of operations/actions that allows to reach a desired goal.
Lately, such approach has been call into question, as the sole focus on the workflow leaves out the \emph{informational context} in which the workflow is executed.

\emph{Artifact-centric models for business processes} recently raised a lot of attention \cite{Bhattacharya2007,Cohn2009}, as they manage to combine structural (i.e. data related) with dynamical (i.e. process related) aspects in a seamless way, thus overcoming the limits of process-centric approach.
In this context, we can see the development of the framework called \emph{Knowledge and Actions Bases} \cite{DBLP:journals/jair/HaririCMGMF13}, the later higher formalization of it named \emph{Description Logic Based Dynamic Systems} \cite{Calvanese2013a}, and the Golog-based work of \cite{Baader2013}.
These works all share the same concept: handle the data-layer through a \emph{Description Logic ontology}, while the process-layer, since DLs are only able to give a static representation of the domain of interest, is defined as \emph{actions} that update the ontology (the so-called ``functional view of knowledge bases'' \cite{DBLP:journals/ai/Levesque84}).
The combination of these two elements generates a transition system in which states are represented by DL knowledge bases.
They do also share a similar objective: verification of temporal formulas over the afore-mentioned transition system.
Since finding a path that lead to a goal state can be expressed as a reachability temporal formula,
these environments can be used for planning purposes, but they are not explicitly meant for this task.
From their definition, we are limited to explore the state-space in a forward manner (we could end up having to explore the full state-space) and only by using the full body of the available knowledge, which is not ideal for developing different ways to search the state-space, as well as under a performance point of view.

In this paper we propose an artifact-centric framework, called \emph{Dynamic Knowledge Bases}, aimed at describing data-rich business domains and be a more versatile environment for planning and decision-making:
%It takes inspiration from the afore-mentioned \emph{Knowledge and Actions Bases}:
the data-layer is taken care of by a DL knowledge base, while a set of actions allows the system to evolve by adding/removing assertions, as well as introducing new instances to the system.
To reach our goals, and overcome the afore-mentioned limitations, our framework relies on few optimizations. 
First of all, although our framework is based on Description Logic, it is desirable to skip completely the use of the TBox: this would allow us to avoid executing reasoning tasks and only work with facts from the ABox, simplifying especially the transition-building process.
We fulfil this aspect with \emph{action rewriting}, which rewrites actions and introduces a \emph{blocking query}:
such query (which is fixed for each action) tells if, given a state, we can perform the given action and built the ending state of the transition, or if the action will lead us to an inconsistent state w.r.t. the TBox.
These operations are done without calculating the ending state, and without the need of the TBox (while keeping the consistency w.r.t. it).

Secondly, while the totality of the available knowledge is necessary to asses the consistency of the overall system, it bounds us to work with details that might not be of interests immediately.
In decision making \cite{Gigerenzer2011}, ``\emph{an heuristic is a strategy that ignores part of the information, with the goal of making decisions more quickly, frugally, and/or accurately than more complex methods}''.
Being able to work with partial information is vital when we deal with systems described by complex ontologies and are composed of millions (if not more) instances.
To allow our framework to be used for such strategies we introduce \emph{partialization}, so that users can focus on a chosen subset of knowledge (partial knowledge);
it allows to build a transition system which starts from a subset of the original ABox (the facts that describe the complete system), and, for each transition, choose which knowledge to transfer to the next state.
Lastly, we demonstrate how, given a path found over the partial knowledge transition system, we can calculate a \emph{global blocking query}, which tells if such path can be performed in the original transition system with no modifications.

The resulting framework constitutes a sound base on top of which researchers can develop new planning techniques useful for all those situations in which is necessary to manipulate both actions and data together (e.g. the decision making process in agents, composition of web services, etc.).

%il framework offre: \\
%- azioni in stile STRIPS (simili, ma non uguali); \\
%- rimozione dell'uso della TBox in tutti i passaggi, lasciando comunque la garanzia della consistenza rispetto ad essa; \\
%- possibilità di lavorare con conoscenza parziale, scegliendo sia uno stato inziale subset di quello originario, sia la conoscenza trasportata da uno stato all'altro in ogni transazione; \\
%- blocking query globale. Dato un piano trovato usato conoscenza parziale, ci dice se è possibile eseguirlo partendo dallo stato iniziale completo, eseguendo esattamente la stessa sequenza di azioni. Dato un piano (ciò dipende dall'algoritmo di planning), la blocking query è sempre calcolabile.
%
%Nel nostro framework, le azioni hanno la possibilità inoltre di introdurre nuovi elementi, quindi i ts generabili possono essere infiniti.
%Noi però non affrontiamo il problema del planning e di trovare un piano finito, lavoriamo solo sul framework che sta alla base, evitando quindi di limitare le possibilità di esso.

% -----------------
\section{Dynamic Knowledge Bases}
\label{sec: Dynamic Knowledge Base}

\emph{Dynamic Knowledge Bases} (DKBs) are, briefly, a variation of \emph{Knowledge and Action Bases} (KABs) \cite{DBLP:journals/jair/HaririCMGMF13}, namely \emph{dynamic systems} (more precisely labelled transition systems) in which states are constituted by DL \emph{knowledge bases} (KBs), and a set of \emph{actions} that makes the system evolve by modifying those KBs.

\begin{definition}
A \emph{DKB} is a tuple $\mathcal{D} = (T, A_0, \Gamma)$, where $(T, A_0)$ is a \textit{DL-Lite}$_\mathcal{A}$ KB, while $\Gamma$ is a finite set of actions.
\end{definition}

We adopt a restricted version of DL-Lite$_\mathcal{A}$ knowledge bases \cite{Calvanese2006},
which does not use attributes (available in full DL-Lite$_\mathcal{A}$ KBs).
DL-Lite$_\mathcal{A}$ employs the \emph{Unique Name Assumption}, thus equality assertions are not allowed.
We adopt DL-Lite$_\mathcal{A}$ as it is, like other DL-Lite dialects, quite expressive while maintaining decidability, good complexity results, and enjoys the FOL-rewritability property.
In the followings, the set $\adom{A}$ identifies the individual constants in the ABox $A$, which are defined over a countably infinite (object) universe $\Delta$ of individuals (it follows that $\textsc{adom}(A) \subseteq \Delta$).
$\mathcal{A}_T$ denotes the set of all possible consistent ABoxes w.r.t. $T$ that can be constructed using atomic concept and atomic role names in $T$, and individuals in $\Delta$.
%
%A DL-Lite KB $(T,A)$ is constituted by a TBox $T$ and an ABox $A$: the TBox contains axioms about concepts and roles that model the domain of interest (the intensional knowledge), while the ABox contains assertions about the individuals that populate such domain (extensional knowledge) defined over a countably infinite (object) universe $\Delta$ of individuals.
%The TBox axioms are a finite set of assertions (nominals are not allowed, to avoid confusion between intensional and extensional levels).
%The ABox assertions are a finite set of facts, i.e., atomic formulas of the form $\cls{C}(\cons{i})$ and $\rol{R}(\cons{i}, \cons{i'}$, where $\cls{C}$ is an atomic concept name, $\rol{R}$ is an atomic role name, and \cons{i}, \cons{i'} are individual constants in $\Delta$.
%The set $\adom{A}$ identifies the constants in $A$ (it follows that $\textsc{adom}(A) \subseteq \Delta$).
%$\mathcal{A}_T$ denotes the set of all possible consistent ABoxes that can be constructed using concept and role names in $T$, and individuals in $\Delta$.
%
The adopted semantic is the standard one based on first-order interpretations and on the notion of model:
a TBox is satisfiable if admits at least one model,
an ABox $A$ is consistent w.r.t. a TBox $T$ if $(T, A)$ is satisfiable,
and $(T, A)$ logically implies an ABox assertion $\alpha$ (denoted $(T, A) \models \alpha$) if every model of $(T, A)$ is also a model of $\alpha$.
% It is assumed that reasoning tasks like checking KB satisfiability and logical implication are decidable.

%The biggest difference between DKBs and KABs is in the definition of actions and the dynamics of the system which depend on them.
We define an {\em action} as: \\
%from the set $\Gamma$ is of the form: \\
\centerline{\acshort{a}{$q , N$}{$E$}} \\
where $\acname{a}$ is the \emph{action name}, $q$ is a query called {\em action guard}, 
$N$ is a set of variables which are used in an \emph{instance creation function},
and $E$ are the {\em action effects}. \\
The guard $q$ is a standard \emph{conjunctive query} (CQ) of the type $q = \exists \overrightarrow{y}.\textit{conj}(\overrightarrow{x},\overrightarrow{y})$,
where $\textit{conj}(\overrightarrow{x},\overrightarrow{y})$ is a conjunction of atoms using free variables $\overrightarrow{x}$ and existentially quantified variables $\overrightarrow{y}$, no individuals.
Atoms of $q$ uses concepts and roles found in $T$. 
\vars{q} represents the variables in $q$ (i.e., $\overrightarrow{x} \cup \overrightarrow{y}$), while \vars{q}$_{\not \exists}$ (resp., \vars{q}$_{\exists}$) only the set $\overrightarrow{x}$ (resp., $\overrightarrow{y}$). \\
% We can see $q^+$ as a set of \emph{positive non-ground atoms}. \\
The set $N$ contains variables which do not appear in $q$ (i.e., $\vars{q} \cap N = \emptyset$), and which are fed to an assignment function $m$ when the action is executed.
The set $E$ is a set of atomic effects (i.e., atomic non-grounded ABox assertions) which is divided in two subsets: the set $E^-$ of \emph{negative effects}, and the set $E^+$ of \emph{positive effects}.
All atoms of $E^-$ must use variables that are in $\vars{q}_{\not \exists}$, while the atoms of $E^+$ uses variables from the set $\vars{q}_{\not \exists} \cup N$.
All variables are defined over a countably infinite (object) universe $V$ of variables.

%\vtodo{possiamo dire che le mie azioni possono essere trasformate facilmente in azioni di tipo STRIPS?}

\begin{definition}
The transition system $\Upsilon_\mathcal{D}$ is defined as a tuple $(\Delta, T, \Sigma, A_0, \Rightarrow)$, where:
(i) $\Delta$ is the universe of individual constants;
(ii) $T$ is a TBox;
(iii) $\Sigma$ is a set of states, namely ABoxes from the set $\mathcal{A}_T$ ($\Sigma \subseteq \mathcal{A}_T$);
(iv) $A_0$ is the initial state;
(v) $\Rightarrow~ \subseteq \Sigma \times \mathcal{L} \times \Sigma$ is a labelled transition relation between states, where $\mathcal{L} = \Gamma \times \Theta$ is the set of labels containing an action instantiation $\acname{a} \vartheta$, where $\acname{a}$ is an action from $\Gamma$ and $\vartheta$ a variable assignment in $\Theta$ from $V$ to $\Delta$.
\end{definition}

\noindent The \emph{transition system} $\Upsilon_\mathcal{D}$ represent the dynamics of a DKB $\mathcal{D}$.
Given a state $A$ and selected an action \acname{a}, the informal semantic of a transition is:
\begin{enumerate}[noitemsep,nolistsep]
\item extract the certain answers~$\answ(q,T,A)$ of the guard~$q$ from the state $A$;
\item pick \emph{randomly} one tuple from \ans{q,T,A} and use it to initiate the variable assignment $\vartheta_\acname{a}$ for the variables $\vars{\acname{a}}$ (at this point we covered only the free variables in $\vars{q}_{\not \exists}$);
% \item execute the UCQ $Q^{ent(E^-,T)} \vartheta_\acname{a}$
\item choose an assignment for the variables in $N$ and use it to extend $\vartheta_\acname{a}$.
We define an assignment function $m(N,A): N \rightarrow (\Delta \setminus \adom{A})$, which assigns to each variable of $N$ an individual from $\Delta$ which does not appear in $A$;
\item use $\vartheta_\acname{a}$ to instantiate the effects $E$ and calculate $A_{next}$ by applying the instantiated effects to $A$.
\end{enumerate}

\noindent The sets $\Sigma$ and $\Rightarrow$ are thus mutually defined using induction (starting from $A_0$) as the smallest sets satisfying the following property:
for every $A \in \Sigma$ and action $\acname{a} \in \Gamma$, if exists an action instantiation \acname{a}$\vartheta_\acname{a}$ s.t. \\
\centerline{$A_{next} = A \setminus sub(ent(E^-,T) \vartheta_\acname{a}, A) \cup E^+ \vartheta_\acname{a}$} 
% $A_{next} \neq A$ (we do not allow self-loops), 
and $A_{next} \in \mathcal{A}_T$, then $A_{next} \in \Sigma$ and $A \overset{l}{\Rightarrow} A_{next}$, with $l = \acname{a} \vartheta_\acname{a}$.
$\acname{a} \vartheta_\acname{a}$ is called an {\em instantiation} of \acname{a}.

$ent(E^-,T)$ represents a set of atoms derived from $E^-$, which represents all the atoms which entail one or more single negative effects $e^-$ in $E^-$ w.r.t. to the TBox $T$.
We take each single negative effect $e^-$ and, by considering $e^-$ as a CQ composed only by one atom, obtain an UCQ $rew_T(e^-)$ by using the \emph{query reformulation algorithm} \cite[Chapter 5.2]{Calvanese2009}.
Since we consider a single atom at a time, the algorithm produces an UCQ composed only by CQs with a single atom $e^-_{rew}$ in them.
Each atom $e^-_{rew}$ either contains variables found in $e^-$ or, in case of a role term, one of the two variables can be a non-distinguished non-shared variable represented by the symbol `\_' (never both variables).
We add each atom $e^-_{rew}$ to the set $ent(E^-,T)$.
Given $ent(E^-,T)$, we calculate the set $sub(ent(E^-,T) \vartheta_\acname{a}, A)$ in the following way.
For each atom $e^-_{rew}$ in $ent(E^-,T)$, we apply the variable transformation $\vartheta_\acname{a}$ to it (the symbol `\_' remains untouched, as it is not linked to any variable that appears in $\vartheta_\acname{a}$);
we then check if it exists in the ABox $A$ an assertion $\alpha$ such that $e^-_{rew} \vartheta_\acname{a} = \alpha$, assuming that the symbol `\_' can be evaluated equal to any individual ($\_ = ind ~,\forall ind \in \adom{A}$).

For clarity, from now on we will denote the set $sub(ent(E^-,T) \vartheta_\acname{a}, A)$ with $E^-_{sub(\vartheta_\acname{a})}$.
Notice that the set $E^-_{sub(\vartheta_\acname{a})}$ is not uniquely determined, as it depends on the ABox on which it is applied.
This behaviour is intentional, as our aim is to have the certainty that an assertion $e^-$ marked for removal will not appear in the next state nor in the ABox $A_{next}$, nor as an inferable assertion ($\langle T, A_{next} \rangle \not \models e^-$); to reach such goal, we have to remove all possible assertions that entail $e^-$.
The set $ent(E^-,T)$, instead, depends only on $E^-$ and $T$, thus it's constant and can be calculated only one time at the beginning.

As we see from the definition of $A_{next}$, actions modify only ABox assertions:
it follows that the TBox is fixed, while the ABox changes as the system evolves (thus an ABox $A_i$ is sufficient to identify the state \emph{i} of the system).
The transition system $\Upsilon_\mathcal{D}$ clearly can be infinite, as we have the possibility to introduce new constants.
We call a \emph{path} $\pi$ a (possibly infinite) sequence of transitions over $\Upsilon_\mathcal{D}$ that start from $A_0$ ($\pi = A_0 \overset{\acname{a_1} \vartheta_1}{\Rightarrow} ... \overset{\acname{a_n} \vartheta_n}{\Rightarrow} A_n$).

\begin{example}
Consider the DKB $\mathcal{D}$ described by the following elements and which models a simple business scenario:
\begin{itemize}[noitemsep,nolistsep]
\item the TBox $T = \{ \cls{Employee} \sqsubseteq \neg \cls{Product}, \cls{Technician} \sqsubseteq \cls{Employee} \}$;
\item the ABox $A_0 = \{ \cls{Technician}(\cons{t1}), \cls{Product}(\cons{p1}) \}$;
\item the action set $\Gamma$ composed of the following actions: \\
\acshort{create}{$\{ \cls{Employee}(x) \}, \{ y \}$}{$\{\cls{Product}(y)\}^+$} \\
\acshort{fire}{$\{ \cls{Employee}(x) \}$}{$\{\cls{Employee}(x)\}^-$} 
\end{itemize}

\noindent If we consider $A_0$ as the initial state in $\Upsilon_\mathcal{D}$, then a possible transition is $A_0 \overset{\acname{create} \vartheta}{\Rightarrow} A_1$ where:
$\vartheta = \{ x \mapsto \cons{t1}, y \mapsto \cons{p2} \}$ (notice that we introduce a new individual $\cons{p2}$),
and $A_1 = \{ \cls{Technician}(\cons{t1}), \cls{Product}(\cons{p1}), \cls{Product}(\cons{p2}) \}$.

We could also perform the action \acname{fire}, as it exists a proper instantiation of it by using the variable assignment $\vartheta_\acname{fire} = \{ x \mapsto \cons{t1} \}$.
The set $ent(E^-,T)$ for the action \acname{fire} corresponds to the set $\{ \cls{Employee}(x), \cls{Technician}(x) \}$, thus $E^-_{sub(\vartheta_\acname{fire})}$ would be equal to $\{ \cls{Technician}(\cons{t1}) \}$.
Performing the action instantiation would get us to the state $A_2 \{ \cls{Product}(\cons{p1}) \}$, and it's clear that $\langle T, A_2 \rangle \not \models \cls{Employee}(\cons{t1})$.
If we would simply remove the instantiated negative effects in $E^- \vartheta_\acname{fire})$, we wouldn't achieve the same result (as the assertion $\cls{Technician}(\cons{t1})$ would still appear in the final state), as if the action didn't have any effect at all.
\end{example}

% -----------------

\section{Optimizations}
\label{sec: Optimizations}

%\input{tex/optimizations_intro.tex}

%\subsection{Preliminary DL-Lite$_\mathcal{A}$ Theorems}
%\label{subsec: Preliminary DL-Lite Theorems}
%
%\input{tex/dl-lite-theorems.tex}

% -----------------
\subsection{Action Rewriting}
\label{subsec: Action Rewriting}

The first optimization we bring to the framework regards actions, and, more specifically, the guard $q$.
%From the definition of the transition function, we see that to extract the certain answers of the guard from the ABox $A$ we need to consider the TBox $T$.
Using the \emph{query reformulation algorithm} \cite[Chapter 5.2]{Calvanese2009}, we can transform a query $q$ into an UCQ $rew_T(q)$ such that $\ans{q,T,a} = \ans{rew_T(q),\emptyset,A}$.
We then take every action \acname{a}, calculate $rew_T(q)$, and, for every CQ $q^{rew} \in rew_T(q)$, create an action \acshort{a^{rew}}{$q^{rew} , N$}{$E$}
(with $N$ and $E$ taken from \acname{a} without modifications).
These new actions slightly modify the transition function $\Rightarrow$: the guard is now evaluated without using the TBox, and the variable assignment $\vartheta_\acname{a^{rew}}$ must be taken from the certain answers \ans{q^{rew},\emptyset,A}, while the rest of the transition function remains the same.

The second optimization regards the ending state of the transition: in the specification of a DKB,
actions could lead to inconsistent states. % w.r.t. the TBox $T$.
We introduce an additional element called \emph{blocking query} $B$, a boolean UCQ used as a block test in the state $A$ before performing the action:
if $B$ returns \emph{false}, then we can perform the action and have the guarantee that the ending state $A_{next}$ is consistent w.r.t. $T$.
The building of $B$ is based on the \emph{NI-closure of $T$} (denoted $cln(T)$) defined in \cite{Calvanese2009} (for the definition of $cln(T)$ we refer the reader to the Appendix).
Each positive effect $e^+ \in E^+$ (column 1 in Table \ref{table: CQs for B}, we need to change the variables accordingly to the ones in $e^+$) could take part in a negative inclusion assertion $\alpha \in cln(T)$ (column 2 in Table \ref{table: CQs for B});
this mean that we have to look for a possible assertion $\beta$ (column 3 in Table \ref{table: CQs for B}) which could break $\alpha$ when $e^+$ is added (\textbf{z} represents a newly introduced variable, thus $\textbf{z} \not \in \vars{q} \cup N \cup \vars{B}$). 
To do so, for each possible $\beta$ we get from $e^+$ and $\alpha$, we perform the following steps (we start from $B = \bot$, where $\bot$ indicates a predicate whose evaluation is \textit{false} in every interpretation):
\begin{enumerate}[noitemsep,nolistsep]
\item \label{blocking query: building the query - queries related to E^+} we check if $\beta$ is present in the positive effects $E^+$ by executing $\ans{\beta, \emptyset, E^+}$ and retrieve all the certain answers $\phi_{E^+}$.
For each $\phi_{E^+}$, it means it exist an assertion $\beta \phi_{E^+}$ which poses a problem.
Since we are dealing with variables (the effects are not instantiated yet), we have to express in $B$ under which conditions $\beta \phi_{E^+}$ would make $A_{next}$ inconsistent;
we do this by adding the corresponding CQ $\beta_{E^+}$ (column 4 in Table \ref{table: CQs for B}) to $B$ by \emph{or}-connecting it to the rest of the CQs.

Notice that we treat \textbf{z} as an existential variable, as it does not appear in $e^+$ and thus we have no constrains about it.

%in the case $\alpha = \textsf{funct} ~\rol{P}$ (or $\alpha = \textsf{funct} ~\rol{P}^-$), the assertion $\beta_1$ used is different, as we need to test the functionality of $\alpha$;

\item \label{blocking query: building the query - queries related to E^-} we check if in $E^-$ there are negative effects that could block $\beta$ by removing it (thus eliminating the threat of an inconsistency).
by executing $\ans{\beta, \emptyset, ent(E^-,T)}$ and retrieve all the certain answers $\vartheta_{E^-}$.
For each $\vartheta_{E^-}$, it means it exist an assertion $\beta \phi_{E^-}$ which is removed.
Since we are dealing with variables (the effects are not instantiated yet), we have to express in $B$ under which conditions $\beta \phi_{E^-}$ can't block an inconsistency in $A_{next}$;
we do this by adding the corresponding UCQ $\beta_{E^-}$ (column 5 in Table \ref{table: CQs for B}) to $B$ by \emph{or}-connecting it to the rest of the CQs.

\item if $E^-$ can't block any inconsistency (thus $\ans{\beta, \emptyset, ent(E^-,T)} = \emptyset$), then we have to express in $B$ under which conditions there will be a inconsistency in $A_{next}$ due to an assertion $\beta$ in $A$ w.r.t $e^+$;
we do so by adding $\beta_A$ (column 6 in Table \ref{table: CQs for B}) to $B$ by \emph{or}-connecting it to the rest of the CQs.
	
\end{enumerate}

Note that while building the blocking query $B$, we could have, for the UCQs $\beta_{E^-}$, inequalities of the type $x \neq \_$, with \_ the non-distinguished non-shared variable generated by $ent(E^-,T)$.
Such inequalities always evaluate to \emph{False}.

%\begin{mdframed}[hidealllines=true]
%\end{mdframed}

\afterpage{%
\clearpage
\begin{landscape}
%\vspace{-10cm}
\begin{table}%[htp]
\makebox[\linewidth]{
\setlength{\tabcolsep}{4pt}
  \begin{tabular}{ l | m{2.2cm} | m{1.5cm} | m{3.4cm} | m{6.5cm} | m{1.8cm} }
    $e^+$		& $\alpha$		& $\beta$ & $\vartheta_{E^+} \rightarrow \beta_{E^+}$ & $\vartheta_{E^-} \rightarrow \beta_{E^-}$ & $\beta_A$ \\ \hline\noalign{\smallskip}
    
    $\cls{A}(x)$	 &
    $\cls{A} \sqsubseteq \neg \cls{A_1}$ \newline $\cls{A_1} \sqsubseteq \neg \cls{A}$ &
    $\cls{A_1}(x)$ &
    $\{ x \mapsto y \} \rightarrow x = y$ &
    $\{ x \mapsto y \} \rightarrow x \neq y \wedge \cls{A_1}(x)$ &
    $\cls{A_1}(x)$ \\ \hline
    
    $\cls{A}(x)$	 &
    $\cls{A} \sqsubseteq \neg \exists \rol{P}$	\newline $\exists \rol{P} \sqsubseteq \neg \cls{A}$ &
    $\rol{P}(x,\textbf{z})$ &
    $\{ x \mapsto y \} \rightarrow x = y$ &
    $\{ x \mapsto y_1, \textbf{z} \mapsto y_2 \} \rightarrow$ \newline $\exists \textbf{z}.\rol{P}(x,\textbf{z}) \wedge x \neq y_1 \wedge \textbf{z} \neq y_2$ &
    $\exists \textbf{z}.\rol{P}(x,\textbf{z})$ \\ \hline
    
    $\cls{A}(x)$	 &
    $\cls{A} \sqsubseteq \neg \exists \rol{P}^-$	\newline $\exists \rol{P}^- \sqsubseteq \neg \cls{A}$ &
    $\rol{P}(\textbf{z},x)$ &
    $\{ x \mapsto y \} \rightarrow x = y$ &
    $\{ x \mapsto y_1, \textbf{z} \mapsto y_2 \} \rightarrow$ \newline $\exists \textbf{z}.\rol{P}(\textbf{z}, x) \wedge x \neq y_1 \wedge \textbf{z} \neq y_2$ &
    $\exists \textbf{z}.\rol{P}(\textbf{z},x)$ \\ \hline
    
    $\rol{P}(x_1,x_2)$	&
    $\exists \rol{P} \sqsubseteq \neg \cls{A}$ \newline $\cls{A} \sqsubseteq \neg \exists \rol{P}$ &
    $\cls{A}(x_1)$ &
    $\{ x_1 \mapsto y \} \rightarrow x_1 = y$ &
    $\{ x_1 \mapsto y \} \rightarrow x_1 \neq y \wedge \cls{A}(x_1)$ &
    $\cls{A}(x_1)$ \\ \hline
    
    $\rol{P}(x_1,x_2)$	&
    $\exists \rol{P}^- \sqsubseteq \neg \cls{A}$ \newline $\cls{A} \sqsubseteq \neg \exists \rol{P}^-$ &
    $\cls{A}(x_2)$ &
    $\{ x_2 \mapsto y \} \rightarrow x_2 = y$ &
    $\{ x_2 \mapsto y \} \rightarrow x_2 \neq y \wedge \cls{A}(x_2)$ &
    $\cls{A}(x_2)$ \\ \hline
    
    $\rol{P}(x_1,x_2)$	&
    $\exists \rol{P} \sqsubseteq \neg \exists \rol{P_1}$ \newline $\exists \rol{P_1} \sqsubseteq \neg \exists \rol{P}$ &
    $\rol{P_1}(x_1, \textbf{z})$ &
    $\{ x_1 \mapsto y \} \rightarrow x_1 = y$ &
    $\{ x_1 \mapsto y_1, \textbf{z} \mapsto y_2 \} \rightarrow$ \newline $\exists \textbf{z}.\rol{P}(x_1,\textbf{z}) \wedge x_1 \neq y_1 \wedge \textbf{z} \neq y_2$ &
    $\exists \textbf{z}.\rol{P_1}(x_1, \textbf{z})$ \\ \hline
    
    $\rol{P}(x_1,x_2)$	&
    $\exists \rol{P}^- \sqsubseteq \neg \exists \rol{P_1}^-$ \newline $\exists \rol{P_1}^- \sqsubseteq \neg \exists \rol{P}^-$ &
    $\rol{P_1}(\textbf{z}, x_2)$ &
    $\{ x_2 \mapsto y \} \rightarrow x_2 = y$ &
    $\{ x_2 \mapsto y_1, \textbf{z} \mapsto y_2 \} \rightarrow$ \newline $\exists \textbf{z}.\rol{P}(\textbf{z}, x_2) \wedge x_2 \neq y_1 \wedge \textbf{z} \neq y_2$ &
    $\exists \textbf{z}.\rol{P_1}(\textbf{z}, x_2)$ \\ \hline
    
    $\rol{P}(x_1,x_2)$	&
    $\exists \rol{P} \sqsubseteq \neg \exists \rol{P_1}^-$ \newline $\exists \rol{P_1}^- \sqsubseteq \neg \exists \rol{P}$ &
    $\rol{P_1}(\textbf{z}, x_1)$ &
    $\{ x_1 \mapsto y \} \rightarrow x_1 = y$ &
    $\{ x_1 \mapsto y_1, \textbf{z} \mapsto y_2 \} \rightarrow$ \newline $\exists \textbf{z}.\rol{P}(\textbf{z}, x_1) \wedge x_1 \neq y_1 \wedge \textbf{z} \neq y_2$ &
    $\exists \textbf{z}.\rol{P_1}(\textbf{z}, x_1)$ \\ \hline
    
    $\rol{P}(x_1,x_2)$	&
    $\exists \rol{P}^- \sqsubseteq \neg \exists \rol{P_1}$ \newline $\exists \rol{P_1} \sqsubseteq \neg \exists \rol{P}^-$ &
    $\rol{P_1}(x_2, \textbf{z})$ &
    $\{ x_2 \mapsto y \} \rightarrow x_2 = y$ &
    $\{ x_2 \mapsto y_1, \textbf{z} \mapsto y_2 \} \rightarrow$ \newline $\exists \textbf{z}.\rol{P}(x_2,\textbf{z}) \wedge x_2 \neq y_1 \wedge \textbf{z} \neq y_2$ &
    $\exists \textbf{z}.\rol{P_1}(x_2, \textbf{z})$ \\ \hline
    
    $\rol{P}(x_1,x_2)$	&
    $\rol{P} \sqsubseteq \neg \rol{P_1}$ \newline $\rol{P_1} \sqsubseteq \neg \rol{P}$ \newline $\rol{P}^- \sqsubseteq \neg \rol{P_1}^-$  \newline $\rol{P_1}^- \sqsubseteq \neg \rol{P}^-$ & $\rol{P_1}(x_1,x_2)$ &
    $\{ x_1 \mapsto y_1, x_2 \mapsto y_2 \}$  \newline $\rightarrow x_1 = y_1 \wedge x_2 = y_2$ &
    $\{ x_1 \mapsto y_1, x_2 \mapsto y_2 \} \rightarrow$  \newline $(x_1 \neq y_1 \wedge \rol{P}(x_1,x_2)) \vee$  \newline $(x_2 \neq y_2 \wedge \rol{P}(x_1,x_2))$ &
    $\rol{P_1}(x_1,x_2)$ \\ \hline
    
    $\rol{P}(x_1,x_2)$	&
    $\rol{P} \sqsubseteq \neg \rol{P_1}^-$ \newline $\rol{P_1}^- \sqsubseteq \neg \rol{P}$ \newline $\rol{P}^- \sqsubseteq \neg \rol{P_1}$  \newline $\rol{P_1} \sqsubseteq \neg \rol{P}^-$ &
    $\rol{P_1}(x_2,x_1)$ &
    $\{ x_1 \mapsto y_1, x_2 \mapsto y_2 \}$  \newline $\rightarrow x_1 = y_1 \wedge x_2 = y_2$ &
    $\{ x_1 \mapsto y_1, x_2 \mapsto y_2 \} \rightarrow$  \newline $(x_1 \neq y_1 \wedge \rol{P}(x_2,x_1)) \vee$  \newline $(x_2 \neq y_2 \wedge \rol{P}(x_2,x_1))$ &
    $\rol{P_1}(x_2,x_1)$ \\ \hline
    
    $\rol{P}(x_1,x_2)$	& 
    $\textsf{funct} ~\rol{P}$ & 
    $\rol{P}(x_1,\textbf{z})$ \newline $\wedge ~x_2 \neq \textbf{z}$ & 
    %$\beta_1 = \rol{P}(x_1,x_2)$ \newline
    $\{ x_1 \mapsto y_1, x_2 \mapsto y_2 \}$  \newline $\rightarrow x_1 = y_1 \wedge x_2 \neq y_2$ & 
    $\{ x_1 \mapsto y_1, \textbf{z} \mapsto y_2 \} \rightarrow$ \newline $(\exists \textbf{z}.\rol{P}(x_1,\textbf{z}) \wedge x_1 \neq y_1 \wedge \textbf{z} \neq x_2) \vee$ \newline $(\exists \textbf{z}.\rol{P}(x_1,\textbf{z}) \wedge x_1 = y_1 \wedge \textbf{z} \neq x_2 \wedge \textbf{z} \neq y_2)$ & 
    $\exists \textbf{z}.\rol{P}(x_1,\textbf{z})$ \newline $\wedge ~x_2 \neq \textbf{z}$ \\ \hline
    
    $\rol{P}(x_1,x_2)$ & 
    $\textsf{funct} ~\rol{P}^-$ & 
    $\rol{P}(\textbf{z},x_2)$ \newline $\wedge ~x_1 \neq \textbf{z}$ & 
    %$\beta_1 = \rol{P}(x_1,x_2)$ \newline 
    $\{ x_1 \mapsto y_1, x_2 \mapsto y_2 \}$  \newline $\rightarrow x_2 = y_2 \wedge x_1 \neq y_1$ & 
    $\{ \textbf{z} \mapsto y_1, x_2 \mapsto y_2 \} \rightarrow$ \newline $(\exists \textbf{z}.\rol{P}(\textbf{z}, x_2) \wedge x_2 \neq y_2 \wedge \textbf{z} \neq x_1) \vee$ \newline $(\exists \textbf{z}.\rol{P}(\textbf{z}, x_2) \wedge x_2 = y_2 \wedge \textbf{z} \neq x_1 \wedge \textbf{z} \neq y_1)$ & 
    $\exists \textbf{z}.\rol{P}(\textbf{z},x_2)$ \newline $\wedge ~x_1 \neq \textbf{z}$ \\ \hline
  \end{tabular}
}
\medskip
\caption{Assertions $\beta_{E^+}$, $\beta_{E^-}$, and $\beta_A$ for a given positive effect $e^+$ and assertion $\alpha$ \label{table: CQs for B}}
%\vspace{-5mm}
\end{table}
\end{landscape}
}

%The symbol $\bot$ indicates a predicate whose evaluation is \textit{false} in every interpretation, while \textbf{z} and $y$ denote, respectively, a newly introduced variable that does not appear in the set $\vars{q} \cup N \cup \vars{B}$ ($\textbf{z} \not \in \vars{q} \cup N \cup \vars{B}$), and a variable that appears in $E^+$ ($y \in \vars{E^+}$).

\begin{definition}\label{def: action rewriting}
Given an action $\acname{a} \in \Gamma$, its \emph{rewritten action} $\acname{a^{rew}}$ is defined as: \\
\centerline{\acshort{a^{rew}}{$q^{rew} , N , B$}{$E$}}
where $q^{rew} \in rew_T(q)$, and $B$ is the \emph{blocking query} of $\acname{a^{rew}}$.
\end{definition}

\noindent The union of all possible rewritten actions defines the set of actions $\Gamma^{rew}$.

\begin{example}
Let's consider the action \acshort{create}{$\{ \cls{Employee}(x) \}, \{ y \}$}{$\{\cls{Product}(y)\}^+$}.
First we calculate $rew_T(q)$, which is the UCQ $\cls{Employee}(x) \vee \cls{Technician}(x)$.

\noindent We can now calculate the blocking query $B$.
We see that the concept term $\cls{Product}$ of the positive effect $e^+ = \cls{Product}(y)$ takes part in the negative-inclusion assertion $\cls{Employee} \sqsubseteq \neg \cls{Product}$, and, by the definition of $cln(T)$, also in the assertion $\cls{Technician} \sqsubseteq \neg \cls{Product}$:
we thus have two $\beta$ assertions, $\cls{Employee}(y)$, and $\cls{Technician}(y)$.
By following the procedure for building $B$, we have no $\beta_{E^+}$ elements (as $\ans{\beta, \emptyset, E^+} = \emptyset$, and no $\beta_{E^-}$ elements (as $\ans{\beta, \emptyset, ent(E^-,T)} = \emptyset$).
The final query is thus composed only of $\beta_A$ elements, and is\\
\centerline{$B = \cls{Employee}(y) \vee \cls{Technician}(y)$}

\noindent We get the following two rewritten actions:\\
\acshort{create^{rew}_1}{$\{ \cls{Employee}(x) \}, \{ y \}, \{ \cls{Employee}(y) \vee \cls{Technician}(y) \}$}{$\{\cls{Product}(y)\}^+$} \\
\acshort{create^{rew}_2}{$\{ \cls{Technician}(x) \}, \{ y \}, \{ \cls{Employee}(y) \vee \cls{Technician}(y) \}$}{$\{\cls{Product}(y)\}^+$}
\end{example}

%\vtodo{questo teorema non serve, è una proprietà delle FOL. Posso quindi rimuoverlo, ma sarebbe bene trovare il riferimento in un libro di logica.}
%
%\begin{theorem}\label{theorem: sub-abox consistency}
%Given a satisfiable DL-Lite$_\mathcal{A}$ KB $(T,A)$, then also $(T,A')$ with $A' \subseteq A$ is satisfiable.
%\end{theorem}
%
%\begin{proof}
%For the proof of the theorem we remind the reader to the Appendix of the extended version of this paper.
%\vtodo{metti link ad Arxiv}
%\end{proof}

\begin{theorem}\label{theorem: action rewriting}
Given a satisfiable KB $(T,A)$, an action $\acname{a^{rew}} \in \Gamma^{rew}$ such that $\vartheta_\acname{a^{rew}} \in \ans{q^{rew},\emptyset,A}$ and $\ans{B \vartheta_\acname{a^{rew}}, \emptyset, A} = \emptyset$, 
then the ABox $A_{next} = A \setminus E^-_{sub(\vartheta_\acname{a^{rew}})} \cup E^+ \vartheta_\acname{a^{rew}}$ is consistent w.r.t. $T$.
\end{theorem}

\begin{proof}
For the proof of the theorem we refer the reader to the Appendix.
\end{proof}

\noindent For the definition of $q_{unsat(T)}$ and $DB(A)$ we refer the reader to the Appendix.

\begin{lemma}
Given an action $\acname{a^{rew}} \in \Gamma^{rew}$, for every ABox $A$ such that $\vartheta_\acname{a^{rew}} \in \ans{q^{rew},\emptyset,A}$ and $\ans{B \vartheta_\acname{a^{rew}}, \emptyset, A} = \emptyset$, we can always perform the transition $A \overset{\acname{a^{rew}} \vartheta_\acname{a^{rew}}}{\Rightarrow} A_{next}$, with $A_{next} \in \mathcal{A}_T$.
\end{lemma}

%\vtodo{da sistemare. Link ad esempio \ref{ex: rewritten action}}

%The UCQ $B$ is intended, as the name implies, to be a block-test: if it returns $\emptyset$ against $A$, then we can perform $\acname{a}$ and be sure to obtain a consistent $A_{next}$ w.r.t. $T$, otherwise we can't perform \acname{a}.
Thanks to the rewriting of actions, we can build the transition system $\Upsilon_\mathcal{D}$ without the need of the TBox $T$, while still having the guarantee that the system is consistent w.r.t. it.

% -----------------
\subsection{Partial Transition System}
\label{subsec: Partial Transition System}

%\vtodo{spiegare meglio i vantaggi della blocking query globale}

We now build a \emph{partialization} $\Upsilon^p_\mathcal{D}$ of the transition system $\Upsilon_\mathcal{D}$, which is built in the same way as $\Upsilon_\mathcal{D}$, apart from two points:
i) the initial state is a subset of the ABox $A_0$
ii) it uses a looser transition function.

\begin{definition}
A \emph{partial transition system} $\Upsilon^p_\mathcal{D}$ is a tuple $(\Delta, T, \Sigma^p, A^p_0, \rightarrow)$, where:
(i) $\Delta$ is the universe of individual constants;
(ii) $T$ is a TBox;
(iii) $\Sigma^p$ is a set of states, namely ABoxes from the set $\mathcal{A}_T$ ($\Sigma^p \subseteq \mathcal{A}_T$);
(iv) $A^p_0$ is a subset of the initial ABox $A_0$ ($A^p_0 \subseteq A_0$);
(v) $\rightarrow~ \subseteq \Sigma^p \times \mathcal{L} \times \Sigma^p$ is a labelled transition relation between states, where $\mathcal{L} = \Gamma^{rew} \times \Theta$ is the set of labels containing an action instantiation $\acname{a^{rew}} \vartheta$, where $\acname{a^{rew}}$ is an action from $\Gamma^{rew}$ and $\vartheta$ a variable assignment in $\Theta$ from $V$ to $\Delta$.
\end{definition}

%Thanks to Theorem \ref{theorem: sub-abox consistency}
As $A^p_0 \subseteq A_0$, we have the guarantee that $A^p_0 \in \mathcal{A}_T$.
The sets $\Sigma^p$ and $\rightarrow$ are mutually defined using induction (starting from $A^p_0$) as the smallest sets satisfying the following property:
for every $A^p \in \Sigma^p$ and action $\acname{a^{rew}} \in \Gamma^{rew}$, if exists an action instantiation $\acname{a^{rew}} \vartheta_\acname{a^{rew}}$ s.t. \\
\centerline{$A^p_{next} \subseteq A^p \setminus E^-_{sub(\vartheta_\acname{a^{rew}})} \cup E^+ \vartheta_\acname{a^{rew}}$} 
and $A^p_{next} \in \mathcal{A}_T$, then $A^p_{next} \in \Sigma^p$ and $A^p \overset{l}{\Rightarrow} A^p_{next}$, with $l = \acname{a^{rew}} \vartheta_\acname{a^{rew}}$.

Notice that $A^p_{next}$ can be any subset of $A^p \setminus E^-_{sub(\vartheta_\acname{a^{rew}})} \cup E^+ \vartheta_\acname{a^{rew}}$, thus allowing to select which knowledge to focus on, unlike in $\Upsilon_\mathcal{D}$ where we transfer all the knowledge from one state to another.
We now define the existing relation between the the partial transition system $\Upsilon^p_\mathcal{D}$ and the transition system $\Upsilon_\mathcal{D}$.
Given a path $\pi^p$ in $\Upsilon^p_\mathcal{D}$, we say that $\pi^p$ is a \emph{proper partialization} of a path $\pi$ in $\Upsilon_\mathcal{D}$ (resp., $\pi$ is a \emph{proper completion} of $\pi^p$) if:
\begin{itemize}[noitemsep,nolistsep,leftmargin=*]
\item each state $A^p_i$ is a subset of the relative state $A_i$ ($A^p_i \subseteq A_i$);

\item each transition is caused by the same action $\acname{a^{rew}_i}$ and the related variable assignments are equal ($\vartheta^p_i = \vartheta_i$).
\end{itemize}

%\vtodo{esempio di proper partialization.}

Between $\Upsilon_\mathcal{D}$ and $\Upsilon^p_\mathcal{D}$ there is no relation such as bisimulation or even simulation;
this is a clear (and intended) consequence of working with partial knowledge.
This also means that we have no immediate way to know if, given a partial path $\pi^p$ in $\Upsilon^p_\mathcal{D}$, we can use the same actions instantiations in $\Upsilon_\mathcal{D}$, and thus if it exists a path $\pi$ that is a proper completion $\pi^p$.
To overcome this problem, we extend the definition of the blocking query $B$ by creating a \emph{global blocking query $B_{\pi^p}$ w.r.t to a finite partial path $\pi^p$}.
$B_{\pi^p}$ is a boolean UCQ that can be evaluated in the complete initial state $A_0$, and, if it is evaluated \emph{False}, gives us the certainty that we can use the same actions instantiations found in $\pi^p$ starting from $A_0$ without generating any inconsistent state w.r.t. $T$.

$B_{\pi^p}$ is built by iteratively adding the single instantiated blocking queries $B_i \vartheta^p_i$ of the actions that compose $\pi^p$ (Algorithm \ref{alg: Global blocking Query},
the symbol $\top$ indicates a predicate whose evaluation is \textit{true} in every interpretation).
At each step, before adding the i-th instantiated blocking query $B_i \vartheta^p_i$ to $B_{\pi^p}$, we perform the following operations:
\begin{itemize}[noitemsep,nolistsep,leftmargin=*]
\item check that $\ans{B_{\pi^p}, \emptyset, E^+_i \vartheta^p_i}$ is \emph{False};

\item remove any CQ $\beta$ in $B_{\pi^p}$ that evaluates always \emph{False} (i.e., contains (in)equalities that evaluates always to \emph{False}, like $\cons{ind_i} = \cons{ind_l}$, or $\cons{ind_i} \neq \cons{ind_i}$);

\item remove from each CQ the (in)equalities that evaluates always to \emph{True}, as they do not influence the ending result.
We are sure that no CQ will be left empty, because it would mean the whole CQ would always evaluate to \emph{True}, and this would have blocked the first step;

\item for each CQ $\beta$, generate a temporary CQ $\beta_{temp}$ by removing all the (in)equalities and transform existential variables in free ones.
Looking at how the blocking query is built, we have that $\beta_{temp}$ is either empty ($\beta$ is composed only of (in)equalities) or contains only one atomic assertion with at most one free variable.
For example, if $\beta = \exists z. \rol{P}(\cons{i_1}, z) \wedge \cons{i_2} \neq z$, then $\beta_{temp} = \rol{P}(\cons{i_1}, z)$;

\item perform $\ans{\beta_{temp}, \emptyset, E^-_{sub(\vartheta^p_i)}}$:
	\begin{itemize}
	\item if it evaluates to \emph{True}, then it means that the instantiated negative effects $E^-_{sub(\vartheta^p_i)}$ remove the atom $\beta_{temp}$, and in this case we can remove the CQ $\beta$ from $B_{\pi^p}$;
	
	\item if it returns answers of the type $\vartheta_{\beta_{temp}} = \{ \textbf{z} \mapsto \cons{ind} \}$, then it means that the instantiated negative effects $E^-_{sub(\vartheta^p_i)}$ remove the atom $\beta_{temp}$ only if $z$ is mapped to the individual $\cons{ind}$.
	We thus add to the CQ $\beta$ the inequality $z \neq \cons{ind}$.
	\end{itemize}
\end{itemize}

\begin{algorithm}[t]
\small
\DontPrintSemicolon
\SetKwInOut{Input}{input}
\SetKwInOut{Output}{output}
\Input{A partial path $\pi^p$}
\Output{An UCQ $B_{\pi^p}$}
\BlankLine

$B_{\pi^p} \asgn~\{ \bot \}$ \;

$i \asgn$ n. of transitions in $\pi^p$ \tcp*[r]{counter variable}

\While(\tcp*[f]{each cycle refers to transition $A^p_{i-1} \overset{\acname{a_i} \vartheta^p_i}{\rightarrow} A^p_i$}){$i > 0$}{
	
	\If{$\ans{B_{\pi^p}, \emptyset, E^+_i \vartheta^p_i} \neq \emptyset$}{
		$B_{\pi^p} \asgn \top$ \tcp*[r]{inconsistency in the i-th transition}
		\textbf{break} \;
	}
	
	\ForEach{$\beta \in B_{\pi^p}$}{
		\If{$\beta$ contains (in)equalities that are always \textit{False}}{
			$B_{\pi^p} \asgn B_{\pi^p} \setminus \beta$ \tcp*[r]{remove CQs that are always \textit{False}}
		}
		
		remove from $\beta$ (in)equalities that are always \textit{True} \;
		
		$\beta_{temp} \asgn \beta$ without (in)equalities and existential operator \;
		
		\uIf{$\ans{\beta_{temp}, \emptyset, E^-_{sub(\vartheta^p_i)}} = True$}{
			$B_{\pi^p} \asgn B_{\pi^p} \setminus \beta$ \tcp*[r]{$E^-_{sub(\vartheta^p_i)}$ erases the CQ $\beta_{temp}$}
		}
		\ElseIf{$\ans{\beta_{temp}, \emptyset, E^-_{sub(\vartheta^p_i)}} \neq \emptyset$}{
			\ForEach {$\vartheta_{\beta_{temp}} = \{ \textbf{z} \mapsto \cons{ind} \} \in \ans{\beta_{temp}, \emptyset, E^-_{sub(\vartheta^p_i)}}$}{
				$\beta \asgn \beta \wedge \textbf{z} \neq \cons{ind}$ \tcp*[r]{update the CQ $\beta$}
			}
		}
	}
	
	$B_{\pi^p} \asgn B_{\pi^p} \cup B_i \vartheta^p_i$ \tcp*[r]{add the blocking query of action $\acname{a_i}$}
	
	$i \asgn i -1$ \;
}

\caption{The algorithm to build the global blocking query $B_{\pi^p}$\label{alg: Global blocking Query}}
\end{algorithm}

\begin{theorem}\label{theorem: Global blocking Query}
Given a DKB $\mathcal{D}$, a finite partial path $\pi^p$, and its global blocking query $B_{\pi^p}$, if $\ans{B_{\pi^p}, \emptyset , A_0} = \emptyset$, then it exists a concretion $\pi$ of $\pi^p$ such that $\pi \in \Upsilon_\mathcal{D}$.
\end{theorem}

\begin{proof}
For the proof of the theorem we refer the reader to the Appendix.
\end{proof}

\begin{example}
Consider the DKB $\mathcal{D}$ described by the following elements and which models a simple business scenario:
\begin{itemize}[noitemsep,nolistsep]
\item the TBox $T = \{ \cls{Stored} \sqsubseteq \neg \cls{Shipped} \}$;
\item the ABox $A_0 = \{ \cls{Product}(\cons{p1}), \cls{Stored}(\cons{p1}), \cls{Product}(\cons{p2}) \}$;
\item the action set $\Gamma$ composed by the following actions:\\
\acshort{pack}{$\{ \cls{Product}(x) \} $}{$\{\cls{Packed}(x)\}^+$},\\
\acshort{ship}{$\{ \cls{Packed}(x) \}$}{$ \{\cls{Shipped}(x)\}^+ $} \\
which becomes the set $\Gamma^{rew}$ composed of the actions:\\
\acshort{pack^{rew}}{$\{ \cls{Product}(x) \} $}{$\{\cls{Packed}(x)\}^+$},\\
\acshort{ship^{rew}}{$\{ \cls{Packed}(x) \}, \{ \cls{Stored}(x) \}$}{$ \{\cls{Shipped}(x)\}^+ $}
\end{itemize}

\medskip

\noindent At this point, we develop a partial transition system $\widehat{\Upsilon}_\mathcal{D}$ by considering the partial initial state $A^p_0 = \{ \cls{Product}(\cons{p1}) \}$.
We can perform the sequence of transitions $\pi^p = A^p_0 \overset{\acname{pack} \vartheta}{\rightarrow} A^p_1 \overset{\acname{ship} \vartheta}{\rightarrow} A^p_2$, where:
$\vartheta = \{ x \mapsto \cons{p1} \}$, $A^p_1 = \{ \cls{Packed}(\cons{p1}) \}$, and $A^p_2 = \{ \cls{Shipped}(\cons{p1}) \}$.
The global blocking query $B_{\pi^p}$ is $\cls{Stored}(\cons{p1})$, and we see that, if we try to transpose $\pi^p$ in the original ABox $A_0$, we have $\ans{B_{\pi^p}, \emptyset , A_0} \neq \emptyset$, thus meaning that $\pi^p$ doesn't have a proper concretion $\pi$ (indeed if we perform the two actions, we would end up having an inconsistent state $A_2$).

If we would consider instead the partial initial state $A^p_0 = \{ \cls{Product}(\cons{p2}) \}$, instead, we woould be able to find a proper completion of $\pi^p$, as $B_{\pi^p}$ would be $\cls{Stored}(\cons{p2})$ and $\ans{B_{\pi^p}, \emptyset , A_0} = \emptyset$.
\end{example}

Given a finite partial path $\pi^p$ and its global blocking query $B_{\pi^p}$, we have a way to know if we can transform $\pi^p$ into a complete path $\pi$ without actually calculating it, only by performing an UCQ over the initial state $A_0$.
Notice also that this result can be applied to all possible ABoxes, not only $A_0$;
as long as $A^p_0$ is contained in an ABox $A$, and $\ans{B_{\pi^p}, \emptyset , A} = \emptyset$, then it exists a path $\pi$ which starts from $A$ and is a proper concretion of $\pi^p$.

% -----------------
%\subsection{Abstract Transition System}
%\label{subsec: Abstract Transition System}
%
%\input{tex/abstract_dynamic_knowledge_base.tex}
%
%% -----------------
%\subsection{Abstract Transition System Properties}
%\label{subsec: Abstract Transition System Properties}
%
%\input{tex/abstract_ts_properties.tex}

% -----------------

\section{Conclusions}

In this paper we formalize a framework, called Dynamic Knowledge Bases, aimed at modelling the dynamics of artifact-centric business processes.
Such framework is represented by a transition system where states are defined by DL-Lite$_\mathcal{A}$ knowledge bases, and where a set of actions allows the system to evolve by adding or removing assertions, along with the possibility to introduce new instances.
The expressive power and reasoning services of Description Logics are very helpful to describe and manage the domain knowledge, but constitute a difficult environment to deal with when it comes to the dynamics of the processes.
To tackle this problem, we introduce two optimizations, namely action rewriting and the partialization of the transition system related to a Dynamic Knowledge Base:
these optimizations give us a framework where we can work with partial knowledge and where the TBox is not needed, still guaranteeing that the resulting system is consistent with it.
Given a path valid for the partial transition system, we can calculate its global blocking query, and know if it can be transferred to the complete transition system without any change, and without the need to do any other calculation.

Our work does not aim to propose a planning technique, neither try to give a solution w.r.t. the decidability/undecidability problem of plan research in our environment (since it is possible to generate an infinite transition system),
but to create a framework that can be used as a formal domain-independent base to develop planning and decision making techniques for data-rich business domains by taking full advantage of the DL-Lite reasoning power.

We are currently working to further expand this framework in various directions.
Under the theoretical side, we are already developing an abstraction of the transition system, in particular by expressing the needed knowledge by using only queries, which can be then used over the complete transition system.
%Another progression is the expansion of the expressive power of the framework, namely by allowing the use of DL-Lite$_{NU}$ knowledge bases, along with more powerful actions (e.g., allowing the use of ECQs to refine the results of the guard).
Under the practical side, we intend to propose a backward planning algorithm, which takes advantage of the abstract transition system and the possibility to work with partial knowledge to return all plans of interest w.r.t. a goal.

Although further investigation is surely needed, Dynamic Knowledge Bases are a promising framework that can be usefully employed to tackle the problem of planning and decision making in artifact-centric business domains.

%\vtodo{l'astrazione lavora su diversi livelli: prima lavoro solo con un set preciso e dove seleziono solo ciò che mi interessa e vedo dove mi porta, poi espando e considero gli effetti completi e vedo se è ancora fattibile nell'astratto, poi passo al concreto.}
%
%\vtodo{Spiegare meglio che impostiamo un framework che può essere utilizzato per il planning, ma non sviluppiamo qua tale tematica.
%
%L'obiettivo è avere un framework che sfrutti la potenza di ragionamento offerto dalle DL, ma che sia comunque abbastanza versatile per farci planning.
%
%Dire inoltre che si, il TS può essere infinito, ma è una cosa voluta in quanto lasciamo all'utilizzatore decidere se usare azioni che introducono nuove istanze o meno.
%Uno può benissimo imporre dei bound.
%
%spiegare meglio i vantaggi che da la blocking query globale.
%
%Dato un piano $\pi^p$, possiamo applicarlo a qualsiasi ABox $A$ per la quale abbiamo $A^p_0 \subseteq A$ e $\ans{B_{\pi^p}, \emptyset , A} = \emptyset$.
%
%Lo combiniamo inoltre alla possibilità di lavorare con della knowledge parziale, che è quello che viene solitamente fatto negli algoritmi di planning, creando così un ponte tra knowledge representation con DL e planning.
%}

% -----------------

\newpage

\bibliographystyle{splncs03}
\bibliography{dl}%,generic}

% -----------------

\newpage

\section*{Appendix}

\subsection*{Definitions}

We put here the definitions of the following elements used throughout the paper:
the NI-closure of $T$ ($cln(T)$), the minimal model $DB(A)$ of a ABox $A$, and the boolean UCQ $q_{unsat(T)}$.
All the definitions are taken from \cite{Calvanese2009}, and are put here to help the reader.

\subsubsection*{NI-closure of $T$}

Let $T$ be a DL-Lite$_\mathcal{A}$ TBox. The NI-closure of $T$, denoted by $cln(T)$, is the TBox defined inductively as follows:
\begin{enumerate}
\item all functionality assertion in $T$ are also in $cln(T)$;

\item all negative inclusion assertion in $T$ are also in $cln(T)$;

\item if $\cls{B_1} \sqsubseteq \cls{B_2}$ is in $T$ and $\cls{B_2} \sqsubseteq \neg \cls{B_3}$ or $\cls{B_3} \sqsubseteq \neg \cls{B_2}$ are in $cln(T)$, then also $\cls{B_1} \sqsubseteq \neg \cls{B_3}$ is in $cln(T)$;

\item if $\rol{Q_1} \sqsubseteq \rol{Q_2}$ is in $T$ and $\exists \rol{Q_2} \sqsubseteq \neg \cls{B}$ or $\cls{B} \sqsubseteq \neg \exists \rol{Q_2}$ are in $cln(T)$, then also $\exists \rol{Q_1} \sqsubseteq \neg \cls{B}$ is in $cln(T)$;

\item if $\rol{Q_1} \sqsubseteq \rol{Q_2}$ is in $T$ and $\exists \rol{Q_2}^- \sqsubseteq \neg \cls{B}$ or $\cls{B} \sqsubseteq \neg \exists \rol{Q_2}^-$ are in $cln(T)$, then also $\exists \rol{Q_1}^- \sqsubseteq \neg \cls{B}$ is in $cln(T)$;

\item if $\rol{Q_1} \sqsubseteq \rol{Q_2}$ is in $T$ and $\rol{Q_2} \sqsubseteq \neg \rol{Q_3}$ or $\rol{Q_3} \sqsubseteq \neg \rol{Q_2}$ are in $cln(T)$, then also $\rol{Q_1} \sqsubseteq \neg \rol{Q_3}$ is in $cln(T)$;

\item if one of the assertions $\exists \rol{Q} \sqsubseteq \neg \exists \rol{Q}$, $\exists \rol{Q}^- \sqsubseteq \neg \exists \rol{Q}^-$, or $\rol{Q} \sqsubseteq \neg \rol{Q}$ is in $cln(T)$, then all three such assertions are in $cln(T)$.
\end{enumerate}

\subsubsection*{Minimal model $DB(A)$}

Let $A$ be a DL-Lite$_\mathcal{A}$ ABox.
We denote by $DB(A) = \langle \Delta^{DB(A)}, \cdot^{DB(A)} \rangle$ the interpretation defined as follows:
\begin{itemize}
\item $\Delta^{DB(A)}$ is the non-empty set consisting of the union of the set of all object constants occurring in $A$;

\item $\cons{a}^{DB(A)} = \cons{a}$, for each object constant $\cons{a}$;

\item $\cls{A}^{DB(A)} = \{ \cons{a} \vert \cls{A}(\cons{a}) \in A \}$, for each atomic concept $\cls{A}$;

\item $\rol{P}^{DB(A)} = \{ (\cons{a_1},\cons{a_2}) \vert \rol{P}(\cons{a_1},\cons{a_2}) \in A \}$, for each atomic role $\rol{P}$.
\end{itemize}

\noindent The interpretation $DB(A)$ is a minimal model of the ABox $A$.

\subsubsection*{Boolean UCQ $q_{unsat(T)}$}

Verifying whether $DB(A)$ is a model of $\langle cln(T),A \rangle$ can be done by simply evaluating a suitable boolean FOL query, in fact a boolean UCQ with inequalities, over $DB(A)$ itself.
A translation function $\delta$ is defined from assertions in $cln(T)$ to boolean CQs with inequalities, as follows:
\begin{align*}
\delta((\textsf{fuct} ~\rol{P})) &= \exists x, y_1, y_2. \rol{P}(x,y_1) \wedge \rol{P}(x,y_2) \wedge y_1 \neq y_2 \\
\delta((\textsf{fuct} ~\rol{P}^-)) &= \exists x_1, x_2, y. \rol{P}(x_1,y) \wedge \rol{P}(x_2,y) \wedge x_1 \neq x_2 \\
\delta(\cls{B_1} \sqsubseteq \neg \cls{B_2}) &= \exists x. \gamma_1(\cls{B_1}, x) \wedge \gamma_2(\cls{B_2}, x) \\
\delta(\rol{Q_1} \sqsubseteq \neg \rol{Q_2}) &= \exists x,y. \rho(\rol{Q_1}, x,y) \wedge \rho(\rol{Q_2}, x,y)
\end{align*}

\noindent where in the last two equations: \\
$\gamma_i (\cls{B},x) = 
\begin{cases}
\cls{A}(x) & \text{if}~ \cls{B} = \cls{A} \\
\exists y_i. \rol{P}(x,y_i) & \text{if}~ \cls{B} = \exists \rol{P} \\
\exists y_i. \rol{P}(y_i,x) & \text{if}~ \cls{B} = \exists \rol{P}^-
\end{cases}
~~~~~~~~~~~~~~~~~~~\rho(\rol{Q},x,y) = 
\begin{cases}
\rol{P}(x,y) & \text{if}~ \rol{Q} = \rol{P} \\
\rol{P}(y,x) & \text{if}~ \rol{Q} = \rol{P}^-
\end{cases}
$

\medskip

\noindent $q_{unsat(T)}$ is then defined with the following steps:
\begin{enumerate}[noitemsep,nolistsep]
\item $q_{unsat(T)} \asgn \bot$;
\item for each $\alpha \in cln(T)$ do: $q_{unsat(T)} \asgn q_{unsat(T)} \cup \{ \delta(\alpha) \}$.
\end{enumerate}

\medskip

\noindent The symbol $\bot$ indicates a predicate whose evaluation is \emph{false} in every interpretation.

\subsection*{Theorem Proofs}

We put here the proofs of the theorems formulated in the paper.

%\subsubsection{Theorem \ref{theorem: sub-abox consistency} proof}
%
%\begin{proof}
%DL-Lite$_\mathcal{A}$ ontology satisfiability is FOL-Rewritable \cite{Calvanese2009}, meaning we can build a query $q_{unsat(T)}$ which, evaluated over the minimal model $DB(A)$, returns \emph{false} if $(T,A)$ is satisfiable.
%$q_{unsat(T)}$ is a boolean UCQ with inequalities; the CQs that compose $q_{unsat(T)}$ are always composed by at least two atoms regarding either base concepts or basic roles.
%It follows that for $q_{unsat(T)}$ to evaluate \emph{true} in $DB(A)$, there must be a pair of assertions $\alpha_1$ and $\alpha_2$ which satisfy at least one of the CQs in $q_{unsat(T)}$.
%The building of $q_{unsat(T)}$ is based only on the TBox $T$, thus independent from the ABox.
%
%If we consider a subset $A'$ of $A$, then we have that $DB(A') \subseteq DB(A)$, as we do not have equalities assertions in DL-Lite$_\mathcal{A}$.
%If we assume that $q_{unsat(T)}^{DB(A')} \neq \emptyset$ (so that $(T,A')$ is not satisfiable), then, for the precedent considerations, there must be two assertions $\alpha_1$ and $\alpha_2$ which satisfy at least one of the CQs in $q_{unsat(T)}$.
%Since $DB(A') \subseteq DB(A)$ these two assertions must be also in $DB(A)$:
%as we assume $(T,A)$ to be satisfiable (thus $q_{unsat(T)}^{DB(A)} = \emptyset$), we have the absurd case where $\alpha_1$ and $\alpha_2$ both satisfy and not satisfy a CQ in $q_{unsat(T)}$.
%\end{proof}

\subsubsection*{Theorem \ref{theorem: action rewriting} proof}

\begin{proof}
From \cite{Calvanese2009}, we know we can check satisfiability of a DL-Lite$_\mathcal{A}$ KB by evaluating the boolean UCQ with inequalities $q_{unsat(T)}$ over the minimal model $DB(A)$.
This is equivalent to find the possible pairs of assertions $\gamma_1$ and $\gamma_2$ that are answers to one of the CQs in $q_{unsat(T)}$.
Each CQ in $q_{unsat(T)}$ is derived from an assertions $\alpha$ in $cln(T)$,
thus, as $\alpha$ is either a negative inclusion or functionality assertion, every CQ is composed of two atoms (which use either a concept term or a role term), and, in case $\alpha$ is a functionality assertion, an inequality.
All variables are defined as existential ones, as the goal of $q_{unsat(T)}$ is just to check whether there is an inconsistency or not, and not understand exactly which assertions generate it.

If we consider the state $A_{next} = A \setminus E^-_{sub(\vartheta_\acname{a^{rew}})} \cup E^+ \vartheta_\acname{a^{rew}}$, we have that $(T, A \setminus E^-_{sub(\vartheta_\acname{a^{rew}})})$ is satisfiable (as we suppose $(T,A)$ to be satisfiable);
it follows that the source of possible inconsistencies are tuples of the type $(\gamma_1, \gamma_2)$ where $\gamma_1$ is from the set of instantiated positive effects $E^+ \vartheta_\acname{a^{rew}}$, and $\gamma_2$ either from $E^+ \vartheta_\acname{a^{rew}}$ or $A \setminus E^-_{sub(\vartheta_\acname{a^{rew}})}$.

%We thus have to check whether adding an assertion $\gamma_1 \in E^+ \vartheta_\acname{a^{rew}}$ generates an inconsistency when coupled with another assertion $\gamma_2$ w.r.t. an assertion $\alpha \in cln(T)$:
%$\gamma_2$ can be either in $A \setminus E^-_{sub(\vartheta_\acname{a^{rew}})}$ or $E^+ \vartheta_\acname{a^{rew}}$.

The UCQ with inequalities $B$ merges the previous considerations and, for every atomic positive effect $e^+$ and for every CQ in $q_{unsat(T)}$ in which one of the atoms is $e^+$, it generates an UCQ which details all the possible cases in which $e^+$ would generate an inconsistency in $A_{next}$, no matter what variable assignment $\vartheta_\acname{a^{rew}}$ is used.
%
% contains, for every atomic positive effect $e^+$, the CQs that catches all the possible assertions $\gamma_2$ in $A_{next}$ which, paired with $\gamma_1 = e^+ \vartheta_\acname{a^{rew}}$, generate an inconsistency w.r.t. an assertion $\alpha$ in $cln(T)$.
We now proceed by giving an example which shows how, given a positive effect $e^+$, $B$ covers all possible inconsistencies.
The remaining cases follow the same logic and are omitted.

Assume that $e^+ = \rol{P}(x_1,x_2)$, and the assertion $\alpha$ in  $cln(T)$ is $\textsf{funct} ~\rol{P}$; it follows that in $q_{unsat(T)}$ we would have the CQ $\beta = (\exists x_1,x_2,\textbf{z}. ~\rol{P}(x_1,x_2) \wedge \rol{P}(x_1,\textbf{z}) \wedge x_2 \neq \textbf{z})$.
As $e^+$ is fixed, and we want to catch specific assertions, we remove $e^+$ from $\beta$ and make all variables free, thus obtaining the CQ $\beta = (\rol{P}(x_1,\textbf{z}) \wedge x_2 \neq \textbf{z})$ that appears in Table \ref{table: CQs for B}.

We now move to check where inconsistencies in $A_{next}$ could be;
as stated before, we can divide this search in the two sets $E^+ \vartheta_\acname{a^{rew}}$ or $A \setminus E^-_{sub(\vartheta_\acname{a^{rew}})}$.
Concerning the set $E^+ \vartheta_\acname{a^{rew}}$, we want to be independent of the particular variable assignment, meaning we have to check in which cases the choosen effect $e^+$ conflicts with other effects in $E^+$;
to do so, we can perform $\ans{\beta, \emptyset, E^+}$ and retrieve all the certain answers $\vartheta_{E^+}$.
%in this way we know which and how atomic positive effects can be transformed in $\beta$.
Assume that there is $e^+_{\gamma_2} = \rol{P}(y_1,y_2)$ in $E^+$, then $\vartheta_{E^+} = \{ x_1 \mapsto y_1, \textbf{z} \mapsto y_2 \}$ is a valid answer to $\ans{\beta, \emptyset, E^+}$; if $x_1$ and $y_1$ are linked to the same instance through the instantiation $\vartheta_\acname{a^{rew}}$ while $x_2$ and $y_2$ not (e.g., $\vartheta_\acname{a^{rew}} = \{ x_1 \mapsto \cons{i_1}, y_1 \mapsto \cons{i_1}, x_2 \mapsto \cons{i_2}, y_2 \mapsto \cons{i_3}, ... \}$), we are going to have an inconsistency in $A_{next}$.
To block such case, we add the CQ $\beta_{e^+} = (x_1 = y_1 \wedge x_2 \neq y_2)$ (as appears in Table \ref{table: CQs for B}) to $B$.

Given the set $A \setminus E^-_{sub(\vartheta_\acname{a^{rew}})}$, 
we first cover the case in which the action has no negative effects ($E^- = \emptyset$), thus leaving us only with the set $A$.
In this situation, our aim is to catch all the assertions $\gamma_2$ in $A$ that, paired with $e^+ \vartheta_\acname{a^{rew}}$, generate an inconsistency.
We could so by evaluating $\ans{\beta \vartheta_\acname{a^{rew}}, \emptyset, A}$, but this would not be a boolean query as $B$ is intended to be, since $\beta$ contains the variable $\textbf{z}$ which is newly introduced and doesn't appear in $\vartheta_\acname{a^{rew}}$.
This doesn't pose a problem, as we actually do not care what exact individual is linked to $\textbf{z}$, but just its existence as long as it is an answer for the CQ $\beta$;
we thus transform the variable $\textbf{z}$ in $\beta$ in an existential one (i.e. $\exists \textbf{z}. ~\rol{P}(x_1,\textbf{z}) \wedge x_2 \neq \textbf{z}$), obtaining the CQ $\beta_A$ in Table \ref{table: CQs for B}.

If the action has negative effects, instead, it could exist an atomic negative effect $e^- = \rol{P}(y_1,y_2)$ that, through the variable assignment $\vartheta_\acname{a^{rew}}$, erases an assertion $\gamma_2$, thus blocking an inconsistency.
To see if such effect exists, we evaluate the query $\ans{\beta, \emptyset, ent(E^-,T)}$ and retrieve all certain answers $\vartheta_{E^-}$, which are variable-to-variable assignments of the form $\{ x_1 \mapsto y_1, \textbf{z} \mapsto y_2 \}$.
If, through a variable assignment $\vartheta_\acname{a^{rew}}$, $y_1$ is assigned to the same individual as $x_1$, and $y_2$ to the same individual as $\textbf{z}$ ($y_1 \vartheta_\acname{a^{rew}} = x_1 \vartheta_\acname{a^{rew}} \wedge y_2 \vartheta_\acname{a^{rew}} = \textbf{z} \vartheta_\acname{a^{rew}}$), then $e^- \vartheta_\acname{a^{rew}}$ has blocked the inconsistency:
if, instead, $y_1 \vartheta_\acname{a^{rew}} \neq x_1 \vartheta_\acname{a^{rew}}$ or $y_1 \vartheta_\acname{a^{rew}} = x_1 \vartheta_\acname{a^{rew}} \wedge y_2 \vartheta_\acname{a^{rew}} \neq \textbf{z} \vartheta_\acname{a^{rew}}$, then $e^-$ doesn't erase $\gamma_2$.
To capture in $B$ the cases illustrated before, we start from the CQ $\beta_A$ obtained before, build the UCQ $\beta_{E^-} = (\exists \textbf{z}.\rol{P}(x_1,\textbf{z}) \wedge x_1 \neq y_1 \wedge \textbf{z} \neq x_2) \vee (\exists \textbf{z}.\rol{P}(x_1,\textbf{z}) \wedge x_1 = y_1 \wedge \textbf{z} \neq x_2 \wedge \textbf{z} \neq y_2)$, and add it to $B$.

\noindent We can repeat the previous steps for each positive effect $e^+$.

\end{proof}

\subsubsection*{Theorem \ref{theorem: Global blocking Query} proof}

\begin{proof}
From the definition of the transition functions $\Rightarrow$ and $\rightarrow$, we see that the ending state of a transition is defined as the assertions from the initial state (minus the negative effects) plus the positive effects; the difference is that in $\Rightarrow$ we add the totality of the assertions to the ending state, while in $\rightarrow$ we can consider a subset of them.
This observation, clearly, is valid along each step of the paths $\pi$ and $\pi^p$.

Given the i-th transition $A^p_{i-1} \overset{\acname{a^{rew}_i} \vartheta^p_i}{\rightarrow} A^p_i$,
and assuming that all previous partial transitions have a proper completion, we need to test $B_i \vartheta^p_i$ over the complete state $A_{i-1}$ to be able to build $A_{i-1} \overset{\acname{a^{rew}_i} \vartheta_i}{\Rightarrow} A_i$ (with $\vartheta_i = \vartheta^p_i$).
The state $A_{i-1}$, though, can be seen as the union of the state $A_{i-2}$ (minus the assertions $E^-_{sub(\vartheta^p_{i-1})}$ removed by the negative effects of action $\acname{a^{rew}_{i-1}}$) plus the assertions $E^+_{i-1} \vartheta_{i-1}$ added by the positive effects of action $\acname{a^{rew}_{i-1}}$.
By checking that $\ans{B_i \vartheta^p_i, \emptyset, E^+_{i-1} \vartheta^p_{i-1}} = \emptyset$ (we know that $\vartheta^p_{i-1} = \vartheta_{i-1}$), we control if the set $E^+_{i-1} \vartheta_{i-1}$ contains any assertion that would create an inconsistency with the positive effects of $\acname{a^{rew}_i}$.
We can then move and check for inconsistencies in the assertions of the set $A_{i-2} \setminus E^-_{sub(\vartheta^p_{i-1})}$, but, as for the single blocking queries $B$, we have to consider the full state $A_{i-2}$ (as we perform $B_i \vartheta^p_i$ over it) and see if the removal of the assertions $E^-_{sub(\vartheta^p_{i-1})}$ could block any inconsistency.

Differently from the building of the single blocking queries, we cannot start from the single positive effect and get the possible assertions we should block, but we have to extrapolate it from $B_i \vartheta^p_i$.
We notice that the CQs in $B_i \vartheta^p_i$ are either composed only of (in)equalities (the $\beta_{E^+}$-type CQs from Table \ref{table: CQs for B}), or an atomic assertion (e.g. $\rol{P}(x,\textbf{z})$) plus some (in)equalities and possibly an existential variable (the $\beta_A$-type and $\beta_{E^-}$-type CQs from Table \ref{table: CQs for B}).
We thus concentrate on the $\beta_A$-type and $\beta_{E^-}$-type CQs, from which we keep only the atomic assertion $\beta_{temp}$, no (in)equalities, and transforming the existential variable in a free one.

Given $\beta_{temp}$, we have two cases:
either it contains only instances, or at maximum one free variable.
In the first case, we simply check whether $E^-_{sub(\vartheta^p_{i-1})}$ contains exactly the same assertion ($\ans{\beta_{temp}, \emptyset, E^-_{sub(\vartheta^p_i)}} = True$), and, in case, remove from $B_i \vartheta^p_i$ the originating CQ of $\beta$, as we are sure that an eventual inconsistency is blocked by the removal effects.
In the second case, instead, we look which specific cases the removal could block (it's the same reasoning applied in the single blocking queries); for each answer $\vartheta_{\beta_{temp}} = \{ \textbf{z} \mapsto \cons{ind} \}$ in $\ans{\beta_{temp}, \emptyset, E^-_{sub(\vartheta^p_i)}}$ we add the inequality $\textbf{z} \neq \cons{ind}$ to the originating CQ of $\beta$, as we are sure that an eventual inconsistency which uses that individual is blocked by the removal effects.
In all other cases we leave the original CQ untouched.

At this point, with the remaining of $B_i \vartheta^p_i$, we can repeat the reasoning done for $A_{i-1}$ in $A_{i-2}$, until we reach $A_0$.
Once we reach $A_0$ (assuming we didn't find any inconsistency before), we add what is left of $B_i \vartheta^p_i$ to $B_{\pi^p}$.
If $\ans{B_{\pi^p}, \emptyset, A_0} = \emptyset$, then we can conclude that also $\ans{B_i \vartheta^p_i, \emptyset, A_0} = \emptyset$, and that the i-th complete transition $A_{i-1} \overset{\acname{a^{rew}_i} \vartheta_i}{\Rightarrow} A_i$ can be performed.
We can apply this reasoning for each transition in $\pi^p$.
\end{proof}

% -----------------

\end{document}